\documentclass[letterpaper, 10 pt, conference]{ieeeconf}  %

\IEEEoverridecommandlockouts                              %

\usepackage{graphicx}
\usepackage{amsmath}
\usepackage{bbm}
\usepackage{dsfont}
\usepackage{subcaption}

\PassOptionsToPackage{dvipsnames}{xcolor}
\PassOptionsToPackage{table}{xcolor}

\usepackage{amssymb}
\usepackage{booktabs}
\usepackage{makecell}
\usepackage[ruled,noend,linesnumbered]{algorithm2e}
\usepackage{titlesec}
\usepackage{wrapfig}
\usepackage{array}
\usepackage{multirow}
\usepackage{colortbl}
\usepackage{tabularx}
\usepackage{macros/mysymbols}

\usepackage{grffile}
\usepackage{xspace}
\usepackage{needspace}

\usepackage{comment} 
\usepackage{url}
\usepackage{bm}
\usepackage{paralist}

\usepackage{pifont}

\usepackage[belowskip=0pt,aboveskip=5pt,font=small]{caption}
\usepackage[belowskip=0pt,aboveskip=0pt,font=small]{subcaption}
\usepackage{xcolor}
\PassOptionsToPackage{hyphens}{url}
\usepackage[hidelinks]{hyperref}
\usepackage{xfrac}

\newcolumntype{Y}{>{\centering\arraybackslash}X}
\newcolumntype{P}[1]{\begin{center}>{\arraybackslash}p{#1}\end{center}}

\usepackage{etoolbox}
\newcounter{magicrownumbers}
\preto\tabular{\setcounter{magicrownumbers}{0}}
\newcommand\rownumber{\stepcounter{magicrownumbers}\arabic{magicrownumbers})\,}

\newcommand{\redtext}[1]{\textcolor{red}{#1}}

\usepackage{cite}

\begin{document}

\title{\LARGE \bf \datasetname: A Dataset and Benchmark\\
for Open-Vocabulary Object Goal Navigation}
\author{
  Naoki Yokoyama$^{1*}$, Ram Ramrakhya$^{1*}$, Abhishek Das$^{2}$, Dhruv Batra$^{1}$, and Sehoon Ha$^{1}$\\
  \thanks{$^{1}$NY, RR, DB, and SH are with the Georgia Institute of Technology; $^{2}$AD is with Meta
{\tt\footnotesize \{nyokoyama, ram.ramrakhya, dbatra, sehoonha\}@gatech.edu, abhshkdz@meta.com}}
}

\bstctlcite{IEEEexample:BSTcontrol}

\maketitle

\let\svthefootnote\thefootnote
\let\thefootnote\relax\footnote{\llap{\textsuperscript{*}}Denotes equal contribution.}
\addtocounter{footnote}{-1}\let\thefootnote\svthefootnote\

\begin{abstract}
We present the Habitat-Matterport 3D Open Vocabulary Object Goal Navigation dataset (\datasetname), a large-scale benchmark that broadens the scope and semantic range of prior Object Goal Navigation (\objnav) benchmarks.
Leveraging the HM3DSem dataset, \datasetname incorporates over \numtotalinstsrough annotated instances of household objects across \numtotalcats distinct categories, derived from photo-realistic 3D scans of real-world environments.
In contrast to earlier \objnav datasets, which limit goal objects to a predefined set of 6-21 categories, \datasetname facilitates the training and evaluation of models with an open-set of goals defined through free-form language at test-time.
Through this open-vocabulary formulation, \datasetname encourages progress towards learning visuo-semantic navigation behaviors that are capable of searching for any object specified by text in an open-vocabulary manner. %
Additionally, we systematically evaluate and compare several different types of approaches on \datasetname.
We find that \datasetname can be used to train an open-vocabulary \objnav agent that achieves both higher performance and is more robust to localization and actuation noise than the state-of-the-art \objnav approach.
We hope that our benchmark and baseline results will drive interest in developing embodied agents that can navigate real-world spaces to find household objects specified through free-form language, taking a step towards more flexible and human-like semantic visual navigation.
Code and videos available at: \projecturl.

\end{abstract}

\section{Introduction}
Visual navigation to a language-specified object is an essential skill for robot assistants that can aid humans in a variety of tasks in indoor environments, such as ``find my keys on the L-shaped couch".
The interest in developing visual navigation systems has increased in recent years, highlighted by the embodied AI community's establishment of standardized evaluation metrics and benchmarks for numerous navigation tasks~\cite{objectnav_tech_report,yadav2023hm3dsem,wani2020multion}.
Various navigation tasks have been proposed, each defining goals differently – point-goal navigation for 2D coordinates~\cite{savva2019habitat}, object-goal navigation ~\cite{deitke2020robothor,yadav2023hm3dsem}, image-goal navigation ~\cite{zhu2017target,krantz2023navigating,chaplot2020neural}, and language-goal navigation (via referring expressions or step-by-step instructions)~\cite{qi2020reverie,anderson2018vision}.
In this work, we focus on the \objnav task where an agent is initialized in an indoor environment and tasked with navigating to an instance of a specified goal object category (\eg, `couch').
\begin{figure}[t]
    \centering
    \includegraphics[width=1.0\columnwidth]{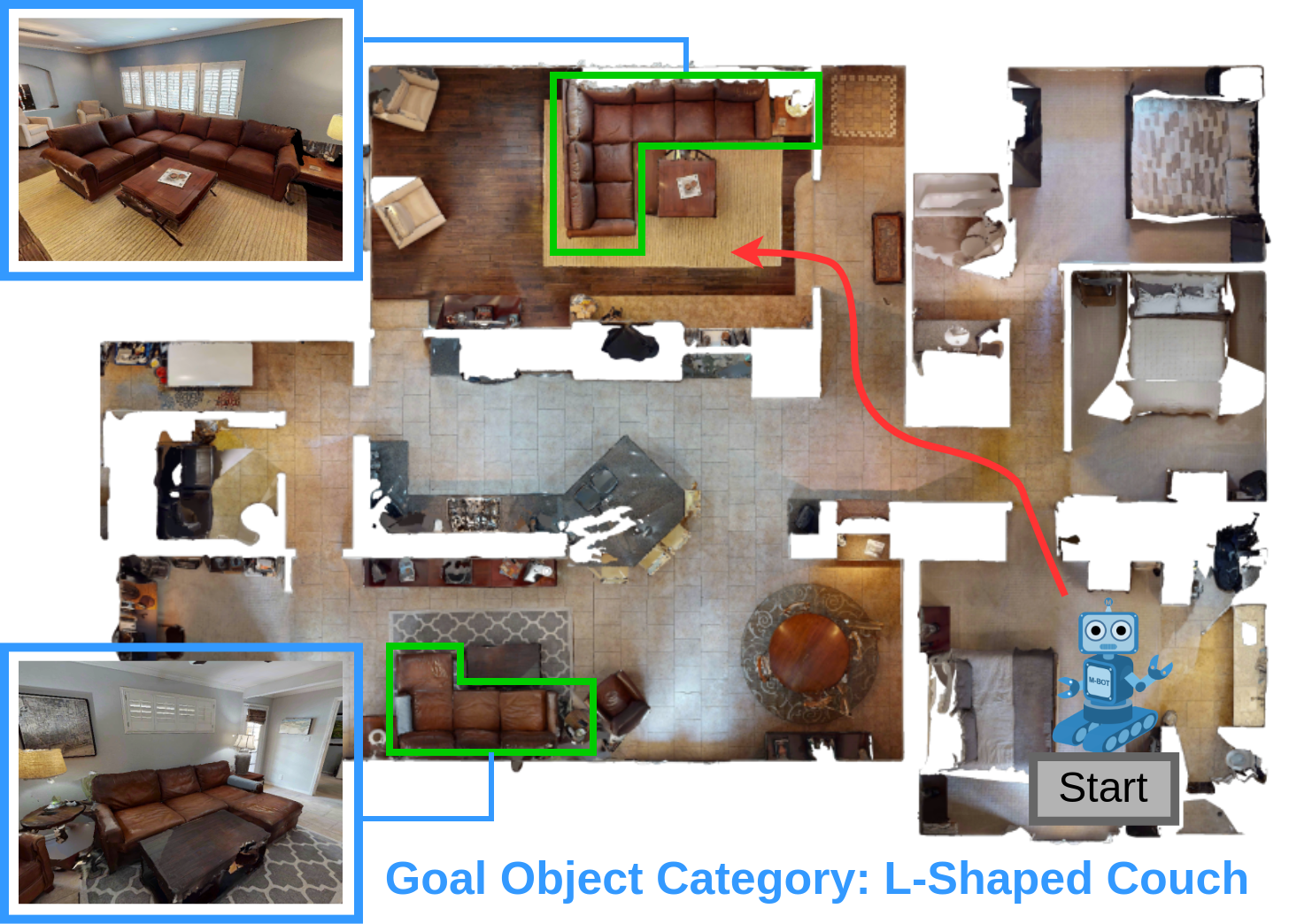}
    \caption{We study the Open-Vocabulary ObjectNav (OVON) task, which involves an agent tasked with navigating to object goals in an open-set, specified through language. In the above example, an agent is tasked with navigating to an `L-Shaped Couch'.}
    \label{fig:teaser}
\end{figure}
While existing \objnav benchmarks have typically concentrated on a limited, fixed set of object categories (6-21 object categories) and have only tested the generalization of navigation agents to novel environments, robotic agents in the real-world must also learn to generalize and navigate to an open set of object goal categories.
To address this, we investigate the problem of open-vocabulary \objnav, where an agent will be asked to navigate to an object specified by language (seen or unseen during training). \figref{fig:teaser} illustrates an example of such an episode.

We introduce a dataset and benchmark named Habitat-Matterport 3D Open-Vocabulary \objnav (\datasetname), designed to test the generalization of \objnav agents in an open-vocabulary setting using the HM3DSem~\cite{yadav2023hm3dsem} scene dataset.
To examine how well agents can generalize to new goal object categories and environments, we propose three evaluation splits:
1) \vseen - consists of goal categories seen during training, 
2) \vue - consists of goal categories synonymous to those seen during training (\ie, ``couch'' category seen during training, evaluated on ``sofa'' during evaluation),
3) \vuh - consists of goal object categories that are not seen during training nor semantically similar to any category from the training set.
We then use these evaluation splits to meticulously examine the performance of a variety of agents across goal object categories with varying degrees of semantic similarity to the training data.

We benchmark policies using several types of popular learning paradigms on \datasetname, including imitation learning (IL), reinforcement learning (RL), and modular methods~\cite{chaplot2020neural,chaplot2020object,yokoyama2024vlfm}, to understand their impact on the agent's ability to navigate to and recognize objects in an open-vocabulary setting.
Through this benchmarking, we find that training end-to-end policies using imitation learning, specifically DAgger~\cite{ross2011dagger}, with frontier exploration trajectories and fine-tuning with RL (referred to as \dagrl) outperforms all other end-to-end trained methods.
However, \dagrl shows a drop in success rate of 11.9$–$23.0\% compared to \vseen when evaluating on \vue and \vuh, suggesting that trained methods struggle to generalize to unseen categories using end-to-end learning on \objnav alone.
In contrast, the modular method VLFM~\cite{yokoyama2024vlfm}, which leverages explicit maps and vision-language foundation models to explore the environment in a semantically meaningful way, achieves consistent performance between 32.4–35.2\% on success rates across the three evaluation splits.
We attribute this to the strong generalization capabilities of the open-vocabulary object detector (OWLv2~\cite{minderer2024scaling}) used to detect the goal object.
Motivated by this observation, we find that augmenting \dagrl with an object detector and a navigation module for bee-lining to the detected object (referred to as \dagrldetect) significantly improves the generalization of end-to-end methods on the \vue and \vuh splits, with a 9.6$–$18.8\% increase in success rate.
We also find that \dagrl is much more robust to noise that simulates real-world conditions than VLFM.

Additionally, we conduct a comprehensive analysis of different architectures used for encoding temporal information (transformer vs. RNN), imitation learning algorithms (behavioral cloning vs. DAgger), and types of trajectories used for imitation learning (frontier exploration vs. shortest path following).
We find that policies perform significantly better when trained using a transformer instead of an RNN, with DAgger instead of behavioral cloning, and with frontier exploration instead of shortest path trajectories.
Our findings on the impact of the types of trajectories used for imitation learning directly contradict the findings presented in SPOC~\cite{spoc2023}, which asserted that shortest path trajectories lead to better performance than those that involve exploration for imitation learning.
Furthermore, we present a detailed analysis of the failure modes of these agents, which illuminates the challenges and opportunities in developing robotic agents capable of robustly navigating to objects specified in free-form language in real-world environments, paving the way for more capable and generalizable visual semantic navigation robots.
Code for \datasetname can be found at \projecturl.

\setlength{\tabcolsep}{6pt}
\begin{table}
    \centering
    \vskip 0.08in
    \resizebox{\columnwidth}{!}{
\begin{tabular}{@{}lccc@{}}
\toprule
& Scene & Object & Object  \\
& type & instances & categories  \\
\midrule
Habitat ObjectNav~\cite{habitatchallenge2023} & Real-world scans & 7,599 & 6 \\
MP3D ObjectNav~\cite{Matterport3D} & Real-world scans & 8,825 & 21 \\
ProcTHOR~\cite{procthor} & Synthetic & 1,633 & 108 \\
OVMM~\cite{homerobotovmm} & Synthetic & 7,892 & 150 \\
\midrule
\datasetname & Real-world scans & \numtotalinsts & \numtotalcats \\
\bottomrule
\end{tabular}
}
    \vspace{5pt}
    \caption{Comparison of public \objnav benchmarks. Our \datasetname benchmark provides a large number of unique objects and categories. \datasetname also uses 3D scans of real-world environments instead of synthetic arrangements of 3D assets that better represent the semantic diversity of real-world conditions.}    
    \label{tab:dataset}
\end{table}

\section{Related Works}

\textbf{\objnav in virtual environments}.
In recent years, several benchmarks have been established for training and evaluating a robot's ability to locate an instance of a given object category within a novel environment (\objnav).
However, these benchmarks often exhibit two main shortcomings.
First, the benchmarks may rely on a limited, fixed set of goal object categories for both training and evaluation.
For instance, the HM3D \objnav dataset~\cite{habitat_challenge2022} encompasses only 6 different goal categories, while the MP3D \objnav dataset~\cite{Matterport3D} includes 21.
This restriction significantly hampers the training of approaches capable of navigating to a broader range of objects, and fails to test an approach's ability to generalize to new goal object categories unseen during training.
Second, benchmarks may exclusively utilize synthetically generated scenes, rather than scans of real-world environments.
For example, ProcTHOR~\cite{procthor} employs procedural generation to create floor plans and populate rooms with 3D assets, while OVMM~\cite{homerobotovmm} utilizes around 200 synthetic 3D scenes designed by humans.
Although synthetic scene creation can produce a vast number of unique environments with less effort than scanning real-world scenes, the quality and realism can be significantly compromised; studies such as \cite{khanna2023hssd} have demonstrated that navigation agents trained on synthetic scenes exhibit poorer generalization to real-world-like environments (e.g., in terms of furniture quantity, types, and arrangement) compared to those trained on fewer, but meticulously designed synthetic scenes by human artists.
To mitigate these issues, our work substantially expands the range of goal object categories, introduces different evaluation splits to assess how well an agent can generalize to new categories, and employs scans of furnished real-world scenes instead of synthetic ones.
Table~\ref{tab:dataset} contrasts our \datasetname dataset with existing public benchmarks.

\textbf{Methods for \objnav}.
Prior works on \objnav falls into two primary categories: modular approaches~\cite{chaplot2020object,ramakrishnan2022poni,yokoyama2024vlfm}, and end-to-end learning via imitation or reinforcement learning~\cite{ye2021auxiliary,thda_iccv21,khandelwal2022,ramrakhya2022habweb}.
Modular methods~\cite{chaplot2020object,ramakrishnan2022poni,yokoyama2024vlfm} break down the \objnav task into sub-skills such as exploration, recognition, and bee-lining (moving to the goal object once detected), employing specific components for each sub-skill.
These approaches utilize heuristic-based exploration strategies, like frontier exploration~\cite{frontier_based_exploration}, supported by explicit spatial and semantic maps, object detection and segmentation models for recognizing goal objects, and path planning algorithms like fast marching methods for waypoint navigation.
End-to-end trained methods leverage neural networks to directly map sensor observations to actions.
Reinforcement learning (RL) variants of these methods learn exploration skills through hand-designed dense rewards~\cite{ye2021auxiliary,thda_iccv21}, while imitation learning (IL) methods draw on extensive human demonstrations~\cite{ramrakhya2022habweb,ramrakhya2023pirlnav} to implicitly learn semantic exploration, or employ shortest path planners~\cite{spoc2023} within procedurally generated environments~\cite{procthor}.
Prior studies employing end-to-end learning often use recurrent neural networks (RNNs) to encode temporal information as the agent navigates its environment~\cite{thda_iccv21,ye2021auxiliary,ramrakhya2022habweb,khandelwal2022,ramrakhya2023pirlnav}.
In contrast, our approach examines the use of transformers to encode observation history for end-to-end methods.
Moreover, we conduct a thorough investigation into the impact of different imitation learning algorithms, types of trajectories used for imitation learning, and architectural choice to encode temporal information, offering an extensive comparison of these methods and guidelines for developing scalable and effective open-vocabulary \objnav agents.

\section{The \datasetname Benchmark}
\label{sec:the-hm3d-ovon-benchmark}

\subsection{\objnav task definition}
\label{subsec:ovon_task}
The \objnav task challenges an agent to locate any instance of a specified goal object category (\eg, `bed') within an unfamiliar environment \cite{objectnav_tech_report}.
At each time step, the agent receives a set of sensory inputs: an RGB image $I_t$, a depth image $D_t$, its relative displacement and heading from the start position (odometry) $P_t=(\Delta x, \Delta y, \Delta \theta)$, and the target object category $G$.
The agent can select one of several actions: \moveforward (by 0.25m), \turnleft and \turnright (by 30$^{\circ}$), \lookup and \lookdown (by 30$^{\circ}$), and \stopac actions.
Success is defined as the agent invoking \stopac within 1m of a goal object within 500 time steps.
In our experiments, we configure the simulated agent to match the specifications of the Stretch robot~\cite{kemp2022design}, which has a height of 1.41m, a base radius of 17cm, and a 360$\times$640 resolution RGB-D camera positioned at a height of 1.31m.

\subsection{The \datasetname dataset}
\label{subsec:open-vocabulary-objectnav-categories}
We utilize the HM3DSem dataset's dense object annotations~\cite{yadav2022habitat} to compile a vast collection of \objnav episodes, termed the \datasetname dataset.
\datasetname includes \numtotalcats goal object categories across \numtotalscenes unique, photorealistic virtual scans of real-world environments.
We ensure goal objects are of significant size and visibility, occupying at least 5\% of the Stretch's camera view from at least one vantage point within 1m of the object to affirm feasibility.
The dataset is segmented into training and evaluation splits, with \numtrainscenes scenes and \numvalscenes scenes, respectively, ensuring no scene or goal object instance overlap between splits.
The training split features goal object instances across \numtraincats categories, whereas the evaluation split comprises \numvalcats categories.

To evaluate generalization to novel objects on varying levels, we divide the evaluation split into three smaller splits, each sharing the same scenes but utilizing mutually exclusive sets of goal object categories:
\begin{itemize}
    \item \vseen: uses goal object categories seen during training.
    \item \vue: uses goal object categories semantically similar to those seen during training (\ie, ``couch'' category seen during training, evaluated on ``sofa'' during evaluation).
    \item \vuh: uses goal object categories semantically divergent from those encountered during training.
\end{itemize}
For \vue and \vuh generation, we first uniformly sample \til25\% of the object categories from \datasetname. Then, we separate these categories using a semantic similarity metric calculated via SentenceBERT~\cite{reimers2019sentencebert}.
SentenceBERT, a fine-tuned variant of the pretrained BERT network, is designed to gauge the semantic similarity between texts by comparing their embeddings' cosine similarity.
We compute SentenceBERT embeddings for each sampled object category and evaluate its cosine similarity with all training split object categories.
An object category is allocated to \vue if it has a maximum similarity surpassing a threshold; otherwise, it is allocated to \vuh.
Table~\ref{tab:val_splits} summarizes the number of categories and the similarity ranges for each split.

\setlength{\tabcolsep}{3pt}
\begin{table}
    \centering
    \vskip 0.08in
    \resizebox{\columnwidth}{!}{
\begin{tabular}{@{}lcccccc@{}}
\toprule
& Novel & Novel goal & Train & Goal \\
& scenes? & object & similarity & object \\
&  & categories? & score range & categories \\
\midrule
\vseen & Yes & No & 1.00 & \numvalseencats \\
\vue & Yes & Yes & [0.68, 0.96] & 50 \\
\vuh & Yes & Yes & [0.45, 0.68] & 49 \\
\bottomrule
\end{tabular}
}
    \vspace{5pt}
    \caption{All evaluation splits in \datasetname use scenes and object instances unseen during training. \vseen uses seen categories, \vue uses unseen categories similar to a seen category, and \vuh uses unseen categories not similar to any seen category. Similarity is determined with SentenceBERT.}
    \label{tab:val_splits}
\end{table}

\subsection{Episode generation}
An episode in \datasetname comprises of a scene, the agent's starting position, and a goal object category.
For episode creation, we first randomly select a goal object category and then randomly determine a starting position adhering to the following criteria:
1) at least one instance of the goal is on the same floor as the starting position, as stair climbing is not anticipated in indoor settings; and
2) the length of the shortest path to the nearest goal location must lie between 1m$-$30m. 
This approach aligns with the episode generation methodology of the \objnav task \cite{habitat_challenge2022}.
\figref{fig:teaser} illustrates a goal example for a single episode.
Following this protocol, we generate 50$k$ episodes per scene for the 145 training scenes, and 3$k$ episodes per scene for the 36 validation scenes.

\section{\datasetname Baselines}

\begin{figure}[t]
    \centering
    \vskip 0.08in
    \includegraphics[width=1.0\columnwidth]{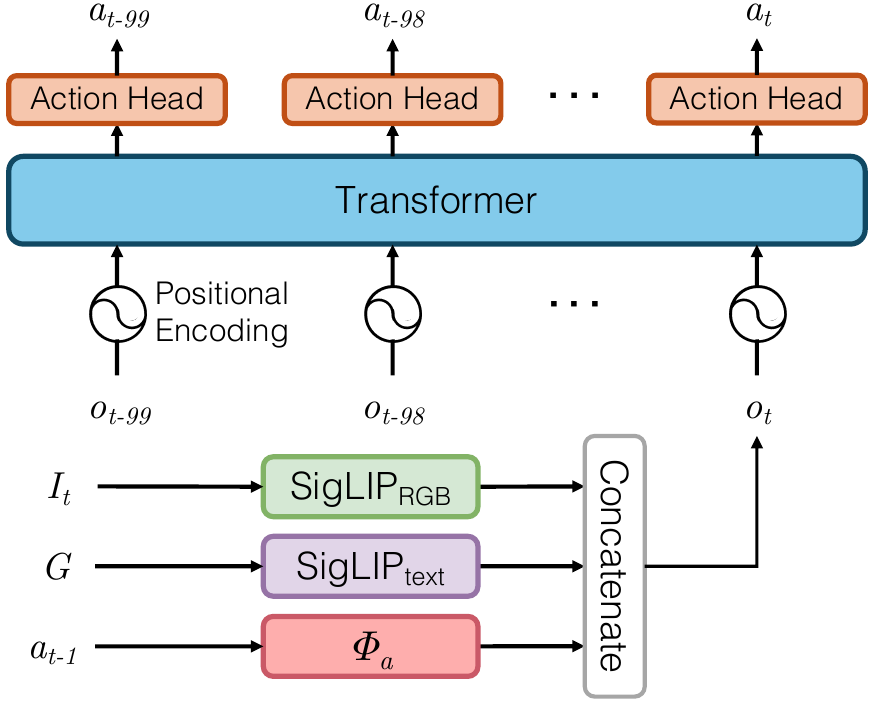}
     \caption{
 Our \ovon policy encodes the current visual observation $I_t$, the goal object category $G$, and the previous action $a_{t-1}$ to form observation embedding $o_t$. At each step, the embedding sequence for the past 100 time steps is fed into a transformer, which uses an action head to sample an action $a_t$.
     }
    \label{fig:architecture}
\end{figure}
In this section, we compare various learning methodologies (imitation learning, reinforcement learning, and modular approaches) and architectural designs (transformer vs. RNN) as proposed in prior studies on object navigation.
We evaluate each method on the \datasetname benchmark.

\textbf{Policy architecture}.
We employ frozen SigLIP~\cite{zhai2023sigmoid} RGB and text encoders to encode the visual observations and the goal object category.
These encoders have been identified as highly effective for \objnav by~\cite{spoc2023}.
The encoders generate two 768-dimensional embeddings for the visual observation, $i_t=\mathrm{SigLIP}_{\mathrm{RGB}}(I_t)$, and the goal object category, $g=\mathrm{SigLIP}_{\mathrm{text}}(G)$.
Additionally, the agent's previous action, $a_{t-1}$, is encoded into a 32-dimensional vector using an embedding layer, $p_t = \phi_a(a_{t-1})$.
These embeddings are concatenated to form the observation embedding, $o_t = [i_t, g_t, p_t]$, which is fed into a 4-layer, decoder-only transformer~\cite{vaswani_nips17} $\pi_{\theta}$ (8 heads, hidden size of 512), with a maximum context length of 100.
$\pi_{\theta}$ takes in the past 100 consecutive observations $[o_{t-99}, ..., o_t]$ and outputs a feature vector for the current time step.
This vector is passed through a linear layer (action head) that predicts a categorical distribution from which an action $a_t$ is sampled, $a_t \sim \pi_{\theta}(\cdot | o_{t-99}, ..., o_t)$.
During RL, an additional linear layer (critic head) is used to project the feature vector into a value estimate for the current state.

When comparing against RNN-based policies, the only architectural change we make is replacing the transformer with a 4-layer LSTM~\cite{hochreiter1997long} with a similar parameter count, for fair comparison.

\begin{figure*}[t!]
\centering
    \vskip 0.08in
    \includegraphics[width=1.0\textwidth, trim=0cm 0cm 0cm 0.1cm, clip]{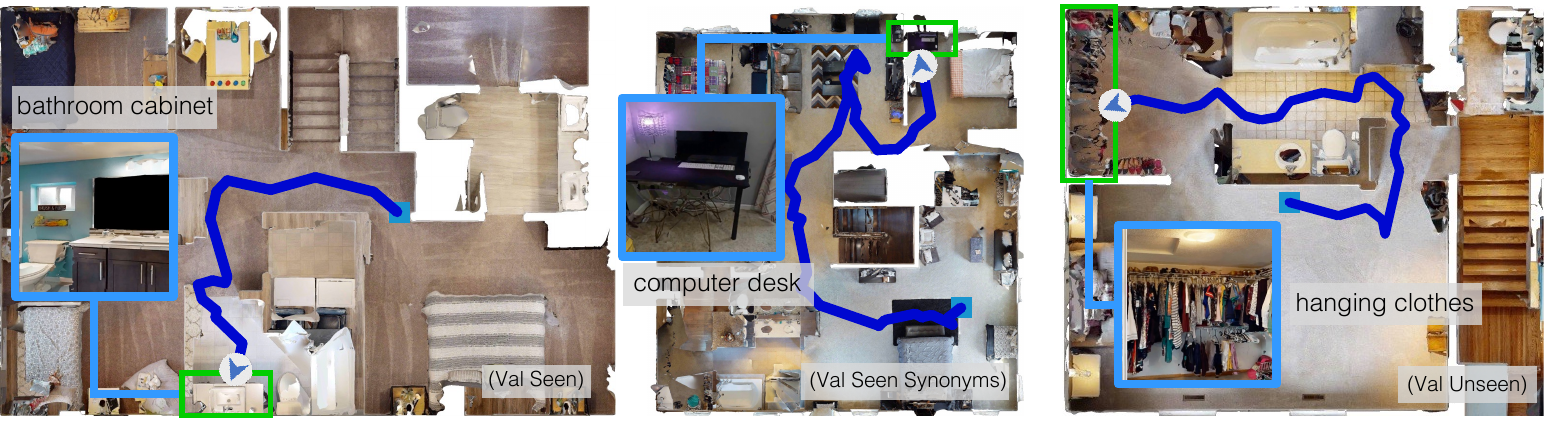}
    \caption{
        Examples of successes of our \dagrl policy for each evaluation split. \dagrl can efficiently explore the environment, avoid obstacles, and stop in front of the goal object when it is spotted, using only RGB observations. Videos can be found at \projecturl.
    }
    \label{fig:sample_trajectories}
\end{figure*}

\textbf{Behavioral cloning (BC).}
Learning from demonstrations has been shown to be a powerful approach for developing efficient semantic navigation behaviors~\cite{ramrakhya2022habweb, yadav2023ovrlv2, ramrakhya2023pirlnav, spoc2023}.
Behavioral cloning employs supervised learning on a dataset of observation-action pairs from expert demonstrations to train policies.
Consider a policy $\pi_{\theta}$ parameterized by $\theta$ that maps observations $o_t$ to an action distribution, $\pi_{\theta}(\cdot|o_t)$.
Let $\tau$ denote a demonstration consisting of observation-action pairs, $\tau = [(o_0,a_0), (o_1, a_1), ..., (o_n, a_n)]$, and $T = \{\tau_i\}_{i}^n$ denote a dataset of demonstrations.
The objective function optimization can be described as:
\begin{equation*}
    \theta^* =
        \text{arg\,max}_\theta
            \sum_{i=1}^N
                \sum_{(o_{t}, a_{t}) \in \tau_i}
                    \log \Big( \pi_{\theta}(a_{t} \mid o_{t}) \Big)
\end{equation*}

While prior work showed that behavioral cloning using demonstrations collected from humans performing the \objnav task can train effective policies~\cite{ramrakhya2022habweb, ramrakhya2023pirlnav}, these demonstrations were limited in diversity and are expensive to collect, especially for the amount of categories in \datasetname.
SPOC~\cite{spoc2023} used a path planner to generate the shortest possible obstacle-free trajectory from the start pose to the goal object, and demonstrated that these trajectories lead to better results than those generated by an expert that exhibits more exploration.
However, this directly contradicts the results of~\cite{ramrakhya2023pirlnav}, which show learning from demonstrations that involve frontier exploration yield better performing policies than shortest path trajectories.
These trajectories are generated by executing frontier-based exploration, which involves the agent systematically moving towards unexplored areas (`frontiers') of the environment, until a goal object is within range (3.5m in our experiments), at which point a shortest path planner is used to plan a path to the goal object (`bee-line').
In this work, we experiment with learning from either frontier exploration or shortest path trajectories.
For behavioral cloning, we generate a trajectory for each episode, for each of the two types (7.25 million trajectories for each type).

\textbf{DAgger.}
DAgger \cite{ross2011dagger} is a supervised learning algorithm that adopts the same loss function as behavioral cloning.
However, unlike behavioral cloning, DAgger involves an expert who provides new action labels for the agent's trajectories `online' during training.
Additionally, the action generated by the policy $\pi_\theta$ is utilized to advance the environment, rather than the expert's actions.
The formulation of DAgger's learning algorithm is similar to behavioral cloning, except each labeled trajectory $\hat{\tau}$ now consists of observation-action pairs $\hat{\tau} = [(o^\pi_0,\hat{a}_0), (o^\pi_1, \hat{a}_1), ...., (o^\pi_n, \hat{a}_n)]$, where action labels $\hat{a}_t$ are provided by the expert, given $o^\pi_t$.

Unlike behavioral cloning, which relies on a pre-recorded dataset, DAgger generates $o^\pi_t$ (and consequently, $\hat{a}_t$) using $a^\pi_{t-1}$ while $\pi_\theta$ is updated, necessitating the capability to interact with the environment and consult the expert online during learning.
Thus, it is crucial to use an expert that can swiftly provide a label $\hat{a}_t$ for $o^\pi_t$, as a slow expert can substantially reduce the speed of training.
Existing implementations of frontier-based trajectory generation for \objnav ~\cite{chaplot2020object, ramrakhya2023pirlnav} are overly slow for in-the-loop execution, taking 250ms per time step, leading to a separation of trajectory generation and supervised learning into discrete stages to maintain training speed.
This constraint has forced prior studies like \cite{ramrakhya2023pirlnav} to adopt behavioral cloning for learning from offline-collected frontier exploration trajectories.
To counteract this, we introduce, alongside our benchmark, an implementation for frontier-based exploration and bee-lining engineered specifically for rapid execution at each time step, ensuring minimal impact on training speed.
Our method only requires an average of 1ms per time step to suggest the appropriate action for guiding the agent from its present position toward the next frontier (or the goal object, if close enough).
This not only facilitates real-time generation of ground-truth expert demonstrations, but also frees the agent from adhering to predetermined trajectories, eliminating the need for a distinct trajectory generation phase prior to training.

\textbf{Reinforcement learning (RL).}
\label{subsec:reinforcement-learning}
To train a policy using RL, we use PPO~\cite{schulman_arxiv17} with a shaped navigation reward commonly used for the PointGoal Navigation task~\cite{wijmans_iclr20}.
This reward consists of $\Delta_{dtg}$ which is the change in the agent's geodesic distance to the nearest goal between time step $t$ and $t-1$, and $r_{success}$ (set to 2.5), the reward received upon success:
\begin{equation*}
r_t =
\begin{cases}
    r_{success} & \text{if success},   \\
    - \Delta_{dtg} - 0.01 & \text{otherwise}   \\
\end{cases}
\end{equation*}

\textbf{\bcrl.}
We pretrain \ovon policies using behavior cloning and fine-tune using reinforcement learning with sparse rewards, following prior work~\cite{ramrakhya2023pirlnav}.

\textbf{\dagrl.}
Similar to \bcrl, we pretrain \ovon policies using DAgger and fine-tune using reinforcement learning with sparse rewards.

\textbf{VLFM.}
Vision-Language Frontier Maps (VLFM)~\cite{yokoyama2024vlfm} is a modular method that achieves state-of-the-art performance on various \objnav benchmarks, and can also support open-vocabulary goal object categories.
VLFM leverages the depth and odometry sensors to build an occupancy map as the robot explores the environment.
In parallel, VLFM utilizes a vision-language foundation model to assess the semantic significance of each explored area in relation to the target object category.
While prioritizing semantically significant areas, VLFM exhaustively explores the environment until an object detector has detected a goal object.
Once a goal object has been detected, the depth camera is used to estimate its coordinates.
The agent then utilizes a policy trained specifically for reaching a given goal coordinate using $D_t$ and $P_t$ to reach the object and call \stopac.
We modify the original VLFM implementation by integrating OWLv2~\cite{minderer2024scaling} as the open-vocabulary object detector, as we found it to outperform GroundingDINO~\cite{liu2023grounding} in our experiments.

\textbf{\dagrldetect.}
This approach extends the \dagrl baseline by incorporating the OWLv2 object detector.
It relies on a trained \dagrl policy to explore the environment until the detector detects a goal object.
Upon detection, the agent transitions from the \dagrl policy to a trained point-goal policy (same as the one used in VLFM) for bee-lining to the detected object.

\textbf{Training details.}
The BC and DAgger policies were trained for 150M steps, as incremental improvements diminished beyond this point.
\bcrl and \dagrl received additional fine-tuning for 150M steps using PPO.
To make comparison with these two fine-tuned policies fair, the RL policy is trained for 300M steps.
All policies were trained across 16 environments per GPU, distributed over 8 TITAN Xp GPUs utilizing Variable Experience Rollout \cite{wijmans2022ver}.

\section{Results}

In this section, we aim to address the following questions:
\begin{enumerate}
\item What differences in performance can be observed between policies trained using different learning methods (RL, IL, or modular learning)?
\item To what extent do various trajectory generation strategies influence the success of imitation learning policies?
\item How do different state encoder architectures (transformers vs. RNNs) impact performance?
\item How robust are our baselines to noise that simulate real-world conditions? 
\end{enumerate}

\subsection{Benchmarking learning methods}
We compare different learning-based approaches: end-to-end policies trained using RL, BC, DAg, \bcrl, and \dagrl, as well as a modular approach, VLFM~\cite{yokoyama2024vlfm}, in Table~\ref{tab:ovon_main}.
We report two metrics – success rate (SR) and Success weighted by Path Length (SPL)~\cite{anderson_arxiv18}, to compare the performance of baselines, across three seeds.
We observe that the policy pretrained using DAgger with frontier exploration trajectories and fine-tuned using RL with sparse rewards, \ie, \dagrl (row 5), outperforms all other end-to-end baselines that do not use a navigation module for bee-lining (rows 1-4).
We attribute this performance to the effective exploration behavior learned using imitation learning from frontier exploration trajectories, and RL fine-tuning that improves the policy's ability to select frontiers more likely to lead to the goal object.
Surprisingly, we find policies trained using RL with a distance to goal reward (row 3) using a transformer state encoder performs comparably to the \dagrl baseline (row 5), only up to 2.1\% worse on SR and up to 2.7\% worse on SPL.
Next, we compare the performance of end-to-end trained methods with a modular, state-of-the-art approach, VLFM~\cite{yokoyama2024vlfm} (row 6). 
While VLFM performs worse than \dagrl on the \vseen split by 6.1\% on SR and 2.6\% on SPL, it outperforms \dagrl on the \vue and \vuh splits by 3.0$-$16.9\% on SR and 2.9$-$11.7\% on SPL, demonstrating strong generalization to unseen object categories.
We find this gap in performance is due to VLFM's use of an open-vocabulary object detector for identifying the goal object.
This detector was trained on a substantially larger set of object classes, which may overlap with classes present in the evaluation splits of \datasetname.
To remedy this issue, \dagrldetect (row 7) is equipped with an open-vocabulary object detector and the same point-goal policy as VLFM for bee-lining to detected goal objects.
We find that \dagrldetect outperforms VLFM on all three evaluation splits by 1.9$-$6.6\% on SR and 0.3$-$4.1\% on SPL.
A characterization of this issue and other failures modes are analyzed in Sec.\ref{sec:failure_analysis}.

\setlength{\tabcolsep}{1.5pt}
\begin{table}
    {\centering
    \vskip 0.08in
    \resizebox{1.01\columnwidth}{!}{
        \begin{tabular}{@{\hspace{-3pt}}clccccccccc}
            \toprule
            & & \multicolumn{2}{c}{\vseen} & \multicolumn{2}{c}{\parbox{2cm}{\centering \textsc{Val Seen} \\ \textsc{Synonyms}}} & \multicolumn{2}{c}{\vuh} \\
            \cmidrule(lr){3-4} \cmidrule(lr){5-6} \cmidrule(lr){7-8}
            & Method & SR$(\mathbf{\uparrow})$ & SPL$(\mathbf{\uparrow})$
              & SR$(\mathbf{\uparrow})$ & SPL$(\mathbf{\uparrow})$ & SR$(\mathbf{\uparrow})$ & SPL$(\mathbf{\uparrow})$ \\
            \midrule
            & 1) BC & 11.1\tiny{$\pm$}0.1 & 4.5\tiny{$\pm$}0.1 & 9.9\tiny{$\pm$}0.4 & 3.8\tiny{$\pm$}0.1 & 5.4\tiny{$\pm$}0.1 & 1.9\tiny{$\pm$}0.2\\
            & 2) DAgger & 18.1\tiny{$\pm$}0.4 & 9.4\tiny{$\pm$}0.3 & 15.0\tiny{$\pm$}0.4 & 7.4\tiny{$\pm$}0.3 & 10.2\tiny{$\pm$}0.5 & 4.7\tiny{$\pm$}0.3\\
            \midrule
            & 3) RL & 39.2\tiny{$\pm$}0.4 & 18.7\tiny{$\pm$}0.2 & 27.8\tiny{$\pm$}0.1 & 11.7\tiny{$\pm$}0.2 & 18.6\tiny{$\pm$}0.3 & 7.5\tiny{$\pm$}0.2\\
            & 4) \bcrl & 20.2\tiny{$\pm$}0.6 & 8.2\tiny{$\pm$}0.4 & 15.2\tiny{$\pm$}0.1 & 5.3\tiny{$\pm$}0.1 & 8.0\tiny{$\pm$}0.2 & 2.8\tiny{$\pm$}0.1\\
            & 5) \dagrl & \textbf{41.3\tiny{$\pm$}0.3} & \textbf{21.2\tiny{$\pm$}0.3} & 29.4\tiny{$\pm$}0.3 & 14.4\tiny{$\pm$}0.1 & 18.3\tiny{$\pm$}0.3 & 7.9\tiny{$\pm$}0.1\\
            \midrule
            & 6) *VLFM \cite{yokoyama2024vlfm} & 35.2 & 18.6 & 32.4 & 17.3 & 35.2 & 19.6\\ %
            & 7) \dagrldetect & 38.5\tiny{$\pm$}0.4 & 21.1\tiny{$\pm$}0.4 & \textbf{39.0\tiny{$\pm$}0.7} & \textbf{21.4\tiny{$\pm$}0.5} & \textbf{37.1\tiny{$\pm$}0.2} & \textbf{19.9\tiny{$\pm$}0.3} \\
            \bottomrule
            \multicolumn{3}{l}{\footnotesize *deterministic method} \\
        \end{tabular}
    }} 
    \caption{Performance of all baselines on the three evaluation splits of \datasetname. \dagrldetect outperforms the state-of-the-art \objnav approach, VLFM.}
    \label{tab:ovon_main}
\end{table}

\subsection{Comparison of trajectories used for imitation learning}
We compare different types of trajectories (frontier exploration vs. shortest path) used during imitation learning.
We find that policies trained using frontier exploration generally outperform those trained using shortest path following, significantly.
Our results are summarized in Table \ref{tab:ovon_experts}; in rows 1, 2, and 4, using frontier exploration rather than shortest path following improves SR and SPL by 4.6$-$6.0\% and 1.0$-$2.2\%, respectively.
For row 3, \bcrl, we found that the poor performance of the pretrained BC policies led to suboptimal sparse RL fine-tuning, resulting in performance that fails to surpass \til20\% in SR for both frontier exploration and shortest path trajectory training.

The observation that trajectories that incorporate exploration are more effective for imitation learning than shortest path trajectories directly contradicts the findings presented by SPOC \cite{spoc2023}.
We attribute this to two possible factors: first, the exploration trajectories in \cite{spoc2023} are different from our frontier exploration trajectories.
While we designate an area of the environment as explored once it has appeared in the agent's field-of-view within a certain distance (3m), the exploration trajectories used in \cite{spoc2023} iteratively visits the next closest room and navigates towards every object within it until at least 75\% of the objects in the room have been seen.
This could consume more steps compared to the more direct method of filling out the map used in frontier exploration, resulting in less efficient learning.
Additionally, its emphasis on objects within a room might not generalize well across different environments, especially if the environments vary widely in layout, room size, or object density.
This contrasts with frontier exploration, which is more focused on spatial layout and could generalize better.
The second possible factor is the difference in the navigation complexity of the environments used. \cite{spoc2023} used synthetic ProcThor~\cite{procthor} scenes, which have been observed to contain much less obstacles and smaller floor plans than real-world scans from HM3D~\cite{khanna2023hssd}.
The higher complexity of HM3D scenes may make policies that learn from trajectories that exhibit exploration behaviors perform better than policies that only learn from shortest paths. 

\begin{figure*}
    \vskip 0.08in
    \centering
    \begin{subfigure}[b]{\textwidth}
        \centering
        \includegraphics[width=\textwidth]{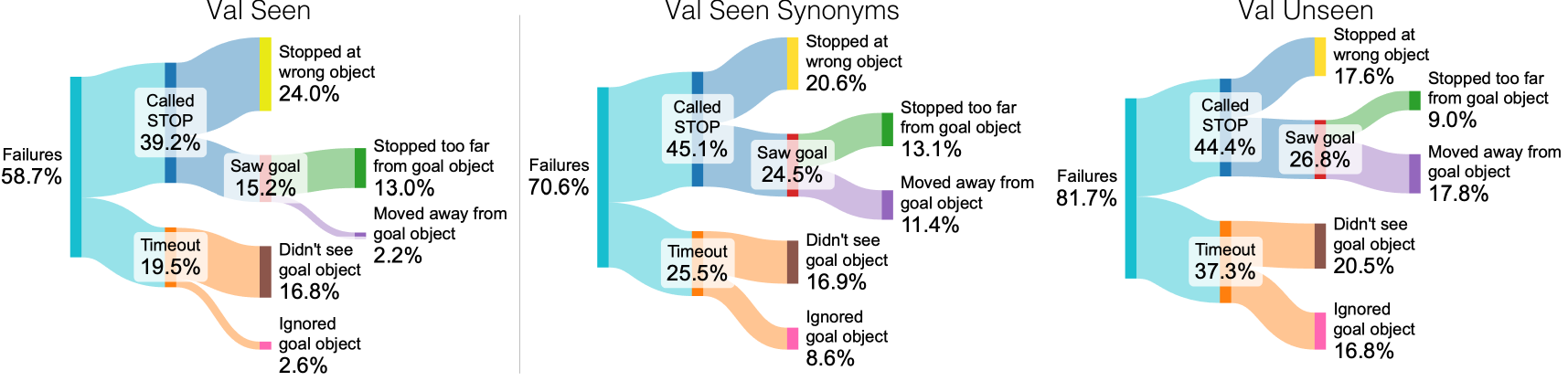}
    \end{subfigure}
    \caption{
       Failure analysis of \dagrl. As the goal object categories become less similar to those seen in training (\ie, \vseen, \vuh), the agent more frequently fails from timeouts (never calling \stopac) and more frequently ignores the goal object.
    }
\label{fig:failure_analysis}
\end{figure*}
\setlength{\tabcolsep}{5pt}
\begin{table}
    \centering
    \vskip 0.08in
    \resizebox{1\linewidth}{!}{
        \begin{tabular}{@{}clccccc@{}}
            \toprule
            & & \multicolumn{2}{c}{\textsc{Shortest Path}} & & \multicolumn{2}{c}{\shortstack{\textsc{Frontier Exploration}}}\\
            \cmidrule{3-4} \cmidrule{6-7}
            \# & Method & SR $(\mathbf{\uparrow})$ & SPL $(\mathbf{\uparrow})$
                & & SR $(\mathbf{\uparrow})$ & SPL $(\mathbf{\uparrow})$ \\
            \midrule
            \\[-10pt]
            \rownumber & BC & 5.1\tiny{$\pm$0.1} & 3.5\tiny{$\pm$0.1} & & \textbf{11.1\tiny{$\pm$0.1}} & \textbf{4.5\tiny{$\pm$0.1}} \\
            \rownumber & DAgger & 13.4\tiny{$\pm$0.2} & 8.0\tiny{$\pm$0.1} & & \textbf{18.1\tiny{$\pm$0.4}} & \textbf{9.4\tiny{$\pm$0.3}} \\
            \rownumber & \bcrl & \textbf{20.6\tiny{$\pm$0.3}} & \textbf{8.5\tiny{$\pm$0.2}} & & 20.2\tiny{$\pm$0.6} & 8.2\tiny{$\pm$0.4} \\
            \rownumber & \dagrl & 37.7\tiny{$\pm$0.6} & 19.0\tiny{$\pm$0.4} & & \textbf{41.3\tiny{$\pm$0.3}} & \textbf{21.2\tiny{$\pm$0.3}} \\
            \bottomrule
            \end{tabular}
    }
    \vspace{5pt}
    \caption{\vseen performance comparing trajectories used for imitation learning. We find policies trained using frontier-exploration perform better than those trained using shortest paths.
    }
    \label{tab:ovon_experts}
\end{table}

\subsection{Transformer vs. RNN}
We examine various techniques for incorporating temporal information from preceding time steps to train effective \ovon policies.
The findings, as presented in Table~\ref{tab:ovon_state_encoder}, demonstrate that transformer-based policies tend to surpass RNN-based counterparts (rows 2, 3, and 5), achieving improvements of up to 6.3\% and 4.5\% in SR and SPL, respectively.
The exceptions, which do not exhibit significant performance enhancement, are the baselines employing behavioral cloning rather than DAgger (rows 1 and 4).
This discrepancy is likely due to the comparative lack of training data diversity in BC when contrasted with DAgger.
DAgger and RL enable the agent to explore previously unencountered trajectories as the policy evolves during training, whereas BC restricts the policy to learning solely from teacher policy trajectories, which remain static throughout the training process.
Given that transformer performance scales more favorably with training data volume, the limited diversity in the training data for behavioral cloning might explain why transformer-based policies did not significantly outperform those based on RNNs.

\setlength{\tabcolsep}{6pt}
\begin{table}
    \centering
    \resizebox{\columnwidth}{!}{
        \begin{tabular}{clccccc}
            \toprule
            & & \multicolumn{2}{c}{\textsc{RNN}} & & \multicolumn{2}{c}{\textsc{Transformer}}  \\
            \cmidrule{3-4} \cmidrule{6-7}
            \# & Method & SR $(\mathbf{\uparrow})$ & SPL $(\mathbf{\uparrow})$
                & & SR $(\mathbf{\uparrow})$ & SPL $(\mathbf{\uparrow})$ \\
            \midrule
            \\[-10pt]
            \rownumber & BC & 10.5\tiny{$\pm$0.3} & 4.1\tiny{$\pm$0.1} & & \textbf{11.1\tiny{$\pm$0.1}} & \textbf{4.5\tiny{$\pm$0.1}} \\
            \rownumber & DAgger & 14.7\tiny{$\pm$0.0} & 7.4\tiny{$\pm$0.2} & & \textbf{18.1\tiny{$\pm$0.4}} & \textbf{9.4\tiny{$\pm$0.3}} \\
            \rownumber & RL & 32.9\tiny{$\pm$0.4} & 15.5\tiny{$\pm$0.2} & & \textbf{39.2\tiny{$\pm$0.4}} & \textbf{18.7\tiny{$\pm$0.2}} \\
            \rownumber & \bcrl & \textbf{21.0\tiny{$\pm$0.4}} & \textbf{11.1\tiny{$\pm$0.2}} & & 20.2\tiny{$\pm$0.6} & 8.2\tiny{$\pm$0.4} \\
            \rownumber & \dagrl & 36.5\tiny{$\pm$0.8} & 16.7\tiny{$\pm$0.3} & & \textbf{41.3\tiny{$\pm$0.3}} & \textbf{21.2\tiny{$\pm$0.3}} \\
            \bottomrule
            \end{tabular}
    }
    \vspace{5pt}
    \caption{Performance on \vseen comparing transformer- and RNN-based architectures for processing temporal information. Transformer-based policy architectures generally perform the best.}
    \label{tab:ovon_state_encoder}
\end{table}

\setlength{\tabcolsep}{2pt}
\begin{table}
    {\centering
    \resizebox{\columnwidth}{!}{
        \begin{tabular}{l@{\hspace{-1pt}}ccccccc}
            \toprule
            \multirow{2}{*}{Method} & \multirow{2}{*}{\shortstack{Eval \\ Noise}} & \multicolumn{2}{c}{\vseen} & 
            \multicolumn{2}{c}{\shortstack{\textsc{Val Seen} \\ \textsc{Synonyms}}}
            & \multicolumn{2}{c}{\vuh} \\
            \cmidrule(lr){3-4} \cmidrule(lr){5-6} \cmidrule(lr){7-8}
             & & SR$(\mathbf{\uparrow})$ & SPL$(\mathbf{\uparrow})$
              & SR$(\mathbf{\uparrow})$ & SPL$(\mathbf{\uparrow})$ & SR$(\mathbf{\uparrow})$ & SPL$(\mathbf{\uparrow})$ \\
            \midrule
            \multirow{2}{*}{VLFM*} & - & 35.2 & 18.6 & 32.4 & 17.3 & 35.2 & 19.6 \\
            & \checkmark & 28.8\tiny{$\pm$0.2} & 14.0\tiny{$\pm$0.1} & 28.5\tiny{$\pm$0.2} & 13.9\tiny{$\pm$0.1} & 29.1\tiny{$\pm$0.3} & 14.2\tiny{$\pm$0.1} \\ 
            \midrule
            \multirow{2}{*}{\dagrldetect} & - & 38.5\tiny{$\pm$0.4} & 21.1\tiny{$\pm$0.4} & 39.0\tiny{$\pm$0.7} & 21.4\tiny{$\pm$0.5} & 37.1\tiny{$\pm$0.2} & 19.9\tiny{$\pm$0.3} \\
            & \checkmark & 38.2\tiny{$\pm$1.0} & 18.1\tiny{$\pm$0.4} & 38.9\tiny{$\pm$0.5} & 18.8\tiny{$\pm$0.4} & 36.9\tiny{$\pm$0.6} & 17.4\tiny{$\pm$0.3} \\
            \bottomrule
            \multicolumn{6}{l}{\footnotesize *deterministic method when evaluated with no noise}
        \end{tabular}
    }} 
    \caption{
    Despite being trained without noise, our \dagrldetect policy is robust against odometry and actuation noise introduced at test-time, since it only relies on visual sensors.
    }
    \label{tab:eval_with_noise}
\end{table}

\subsection{Robustness to noise}
In perceiving the environment, the policies we train for \objnav rely solely on an RGB sensor.
This makes them inherently less sensitive to types of perturbations that may affect other approaches that rely on building maps upon which observations are spatially projected using odometry and depth sensors.
We evaluate our best baseline, \dagrldetect, and VLFM in the presence of noise that simulates real-world conditions.
We use an actuation noise model within Habitat acquired from mocap-based benchmarking recorded using a robot with a dynamics model and sensor suite similar to the Stretch~\cite{pyrobot2019}.
This adds Gaussian noise to the position and rotation of the robot each time it takes an action.
We also add noise uniformly sampled sensor noise using values drawn from the uncertainty study done by \cite{habitatsim2real20ral}, adding between $\pm$7mm and $\pm$2$^{\circ}$ to the odometry sensor $P_t$ at each time step.
Note that neither \dagrldetect nor VLFM use models that were trained in the presence of these types of noise.

The results of evaluation with noise are shown in Table~\ref{tab:eval_with_noise}.
We find that VLFM is sensitive to noise, and drops in SR and SPL by up to 6.1\% and 5.4\%, respectively.
This can be attributed to the fact that VLFM iteratively builds top-down maps that it relies on to decide where to explore next, and the added noise can make the maps innaccurate.
In contrast, \dagrldetect is robust to noise, and drops in SR and SPL by only up to 0.3\% and 3.0\% respectively.
Since the \dagrl policy used for exploration is only able to use RGB observations from a sequence of previous consecutive time steps to determine where it currently is relative to where it has been before, the policy does \textit{not} learn to rely on precise localization to attain good performance.
Thus, it is robust to the additional noise that is injected at test-time.
The larger drop in SPL compared to SR can be attributed to the fact that the noise causes the robot to deviate from its desired path, and the additional corrective actions required leads to longer paths compared to evaluation without noise.

\subsection{Failure analysis}
\label{sec:failure_analysis}

To characterize the types of behavior that can be learned from \datasetname, we present a detailed analysis of the different failure modes of \dagrl, the policy that attains the best performance without relying on an external open-vocabulary object detector.
A breakdown of the different failures modes on each of the three evaluation splits is visualized in \figref{fig:failure_analysis}.
As the goal object categories in the evaluation split decrease in similarity to those seen during training (\vseen$\rightarrow$\vue$\rightarrow$\vuh), the following trends occur:
We observe that failures that occur from timeout (\ie, never calling \stopac) increase (orange in \figref{fig:failure_analysis}, 19.5\%$\rightarrow$25.5\%$\rightarrow$37.3\%), primarily due to ignoring the goal object (pink, 2.6\%$\rightarrow$8.6\%$\rightarrow$16.8\%), rather than inefficient exploration that causes the agent to miss the goal object (brown, stays between 16.8\%-20.5\%).
We also observe that the agent is more likely to move away from the goal object (enter and leave its success region) and call stop elsewhere (purple, 2.2\%$\rightarrow$11.4\%$\rightarrow$17.8\%).
These trends indicate that the agent struggles to generalize and correctly navigate to and call \stopac when seeing a goal object that is too semantically different from the goal categories in \datasetname's training split.
However, as shown by the performance of \dagrldetect, these shortcomings can be addressed using a pretrained open-vocabulary object detector and a way to navigate the robot to a detected object and call \stopac (\ie, a point-goal policy).

The causes of failure that consistently contribute a large portion of the failure across all three of the evaluation splits are: (1) stopping at the wrong object (yellow, between 16.8\%-20.5\%), and (2) not seeing the goal object (brown, between 16.8\%-20.5\%).
This indicates that future work should prioritize reducing false positives and improving exploration to find the goal object within the time step limit.

\section{Conclusion}
We present \datasetname, a large-scale dataset and benchmark that provides \numtotalcats goal object categories and over \numtotalinstsrough annotated instances of household objects across \numtotalscenes unique, photo-realistic
virtual scans of real-world environments.
\datasetname facilitates the training and evaluation of models with an open-set of goals defined through free-form language, compared to previous datasets that are limited to a predefined set of object categories at test-time. Through extensive experiments, we demonstrate that \datasetname can be used to train an open vocabulary \objnav agent that achieves both higher performance and better robustness to localization and actuation noise than the state-of-the-art. We hope that \datasetname leads to further advancements in embodied AI and opens up new avenues for research in visual semantic navigation and object recognition.

\section{Acknowledgements}
\scriptsize{
    The Georgia Tech effort was supported in part by Hyundai Mobis, ONR YIPs, and ARO PECASE. The views and conclusions are those of the authors and should not be interpreted as representing the U.S. Government, or any sponsor.
}

{
    \bibliographystyle{style/IEEEtran}
    \bibliography{bib/strings,bib/main}
}

\end{document}